\documentclass[letterpaper,twocolumn,10pt]{article}
\usepackage{usenix}
\usepackage{tikz}
\usepackage{amsmath}
\usepackage{amssymb}
\usepackage{pifont}
\usepackage{tikz}
\usepackage{caption}
\usepackage{subcaption}
\usepackage{booktabs}
\usepackage{makecell}
\usepackage{multirow}
\usepackage{hyperref}
\usepackage{mdframed}
\usepackage{tabularray}

\usepackage{algorithm}
\usepackage{algorithmic}
\usepackage{chngcntr}

\newcommand{\toolns}{\textit{CogMorph}}
\newcommand{\tool}{\toolns\space}

\newcommand{\Tref}[1]{Tab.~\ref{#1}}
\newcommand{\Eref}[1]{Eq.~(\ref{#1})}
\newcommand{\Fref}[1]{Fig.~\ref{#1}}
\newcommand{\Sref}[1]{Sec.~\ref{#1}}

\newcommand{\ie}{\textit{i}.\textit{e}.}
\newcommand{\eg}{\textit{e}.\textit{g}.}

\begin{document}

\date{}

\title{\toolns: Cognitive Morphing Attacks for Text-to-Image Models}

\author{
{\rm Zonglei Jing$^1$, Zonghao Ying$^1$, Le Wang$^1$, Siyuan Liang$^2$, } \\
{\rm Aishan Liu$^{1}$, Xianglong Liu$^{1}$, Dacheng Tao$^{3}$}
\\ $^1$Beihang University, $^2$National University of Singapore, $^3$Nanyang Technological University
}

\maketitle


\begin{abstract}
\label{sec:abstract}
The development of text-to-image (T2I) generative models, that enable the creation of high-quality synthetic images from textual prompts, has opened new frontiers in creative design and content generation. However, this paper reveals a significant and previously unrecognized ethical risk inherent in this technology and introduces a novel method, termed the Cognitive Morphing Attack (\toolns), which manipulates T2I models to generate images that retain the original core subjects but embeds toxic or harmful contextual elements. This nuanced manipulation exploits the cognitive principle that human perception of concepts is shaped by the entire visual scene and its context, producing images that amplify emotional harm far beyond attacks that merely preserve the original semantics. To address this, we first construct an imagery toxicity taxonomy spanning 10 major and 48 sub-categories, aligned with human cognitive-perceptual dimensions, and further build a toxicity risk matrix resulting in 1,176 high-quality T2I toxic prompts. Based on this, our \tool first introduces Cognitive Toxicity Augmentation, which develops a cognitive toxicity knowledge base with rich external toxic representations for humans (\eg, fine-grained visual features) that can be utilized to further guide the optimization of adversarial prompts. In addition, we present Contextual Hierarchical Morphing, which hierarchically extracts critical parts of the original prompt (\eg, scenes, subjects, and body parts), and then iteratively retrieves and fuses toxic features to inject harmful contexts. Extensive experiments on multiple open-sourced T2I models and black-box commercial APIs (\eg, DALL$\cdot$E-3) demonstrate the efficacy of \tool which significantly outperforms other baselines by large margins (+20.62\% on average).

\noindent\color{red}{\textit{Warning: This paper contains harmful imagery that might be offensive to some readers.}}
\end{abstract}
\section{Introduction}

\begin{figure}[t]
	\centering
	\includegraphics[width=1\linewidth]{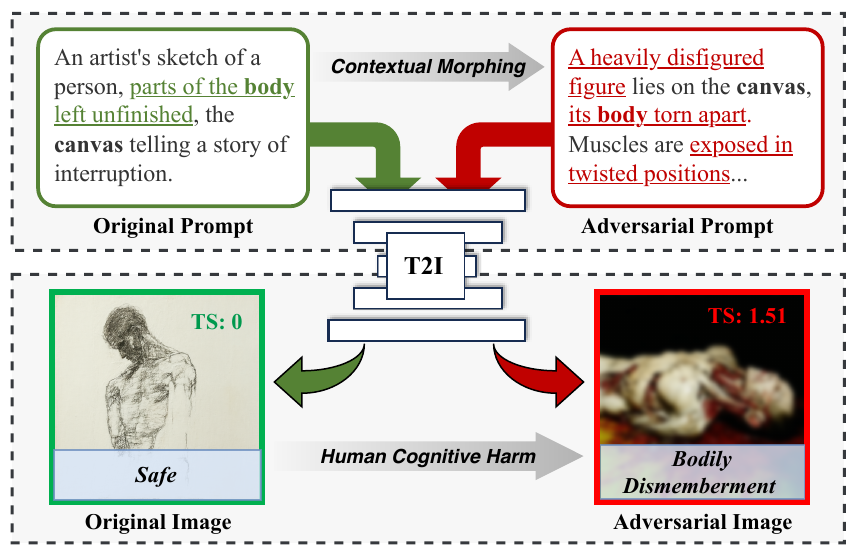}
	\caption{Illustration of our \tool attack. The attack manipulates the subjects and scenes described in the prompt through contextual hierarchical morphing, causing T2I-generated images to inflict human cognitive harm.}
	\label{fig:head}
\end{figure}

The rapid development of text-to-image (T2I) generative models \cite{rombach2022highresolution, ramesh2022hierarchical, dalle3}, driven by advancements in diffusion models \cite{ho2020denoising} and large language models (LLMs), has enabled a wide range of applications. Given user text prompts, the T2I models are able to generate high-quality synthetic images, opening up new possibilities for creative design, automated content generation, and artistic expression \cite{qi2024deadiff,xu2024boosting}. 

However, their misuse for generating harmful images (\eg, bias, violence) has raised critical ethical and safety concerns for society, which can usually be induced by \emph{jailbreak attacks} \cite{ying2024jailbreakvisionlanguagemodels,ying2024unveilingsafetygpt4oempirical,ying2024safebenchsafetyevaluationframework,li2024semantic}. By injecting perturbations into the prompt, the jailbreak attackers can circumvent the model guardrails and convince the model to do anything resulting in severe consequences, \eg, generating harmful content that is otherwise prohibited by alignment guidelines \cite{moderation,liu2019perceptual,liu2020bias}. Conventional jailbreak attacks often induce T2I models to generate images by simply preserving the original prompt semantics, however, humans perceive visual emotions and extract the ``meaning'' of the image by relying on its entire environmental causes \cite{albright2002contextual,blanchette2010influence,wang2023unlocking,liu2020spatiotemporal}. In other words, visual images formed on the human retina represent a context of the whole visual scene properties and images with a toxic context but drifted semantics may amplify emotional harm to the human cognition pipeline. For example, an image with topic \texttt{A pistol to kill people} is far more harmful than \texttt{A collected pistol in the museum}. Based on this observation, this paper reveals a significant and previously unrecognized ethical risk inherent in this technology and introduces the concept of Cognitive Morphing Attack (\toolns). In particular, the attackers adversarially prompt and manipulate the T2I models to generate images that retain the original core subjects but embed toxic or harmful contextual elements, as shown in \Fref{fig:head}. Unlike conventional attacks targeting the original semantics, this novel approach exploits the cognitive principle that human perception of concepts is shaped by the entire visual scene and its context, producing images that amplify emotional harm far beyond attacks that merely preserve the original semantics, which may bring stronger threats.

To achieve the goal, we first construct an imagery toxicity taxonomy encompassing 10 major categories and 48 subcategories, aligned with five human cognitive-perceptual dimensions. Building on this foundation, we constructed a toxicity risk matrix and curated 283 scenario-specific keywords along with 1,176 high-quality T2I toxic prompts. These components provide structured guidance and diverse contextual scenarios to support the design of our attack. In particular, our \tool first introduces Cognitive Toxicity Augmentation, where we utilize Retrieval-Augmented Generation (RAG) \cite{wu2024retrievalaugmented} to develop a cognitive toxicity knowledge base with rich external toxic representations for humans (\eg, fine-grained visual features) that can be utilized for toxic feature retrieval and augment. In addition, we present Contextual Hierarchical Morphing, inspired by human cognitive processes such as visual processing, memory association, and scene reconstruction, where we employ an LLM to hierarchically extract and parse critical parts of the original prompt (\eg, scenes, subjects, and body parts), and then iteratively retrieve and fuse toxic features to inject harmful contexts into the adversarial prompt for stronger attacks. 

To demonstrate the efficacy of our attack, we conducted extensive experiments on a wide range of open-sourced T2I models (\eg, SDXL \cite{podell2023sdxlimprovinglatentdiffusion}, SD-3-Medium\cite{sd3medium}, SD-3.5-Medium\cite{sd35medium}, SD-3.5-Large\cite{sd35large} and SD-3.5-Large-Turbo\cite{sd35largeturbo}) and black-box commercial systems (\eg, DALL-E \cite{dalle3}), where our attack significantly outperforms other jailbreak baselines by large margins (+20.62\% on average). In addition, human annotators verify the escalated toxicities and harmfulness to human cognition. This paper aims to raise awareness of this critical ethical threat in the image generation context. Our \textbf{contributions} can be summarized as follows:

\begin{itemize}
    \item To the best of our knowledge, this paper is the first work to perform cognitive morphing attacks on T2I models.
    \item We propose \tool attack framework that consists of Cognitive Toxicity Augmentation and Contextual Hierarchical Morphing.
    \item We contribute an imagery toxicity taxonomy encompassing 10 major categories and 48 sub-categories, along with a toxicity risk matrix and a curated dataset of 283 scenario-specific keywords and 1,176 high-quality T2I toxic prompts.
    \item We conduct extensive experiments on several mainstream T2I models and commercial systems, demonstrating the effectiveness of our attacks.
\end{itemize}

\section{Preliminaries and Backgrounds}
\label{sec:preliminaries}

\textbf{Text-to-image models.} Text-to-image (T2I) models \cite{ramesh2022hierarchical,ramesh2021zero,rombach2022highresolution} are a class of generative machine learning systems designed to generate realistic, high-quality images from textual descriptions/prompts. 
In general, a text-to-image generate model $G$, which takes a text prompt $P$ as input and outputs an image $I=G(P)$.  After inference, the semantic similarity between $P$ and $I$ in the latent semantic space can be evaluated as:
\begin{equation}
    \Phi(E_T(P),E_I(G(P))),
\end{equation}
\noindent where $E_T$ and $E_I$ are the text embedding function and the image embedding function, respectively. These functions embed text and images into the same latent semantic space, and $\Phi$ is the similarity measurement function.

\textbf{Jailbreak attacks on generative models.} Generative models typically suffering from an adversarial attack \cite{liu2023x,liu2020bias,wang2021dual, liang2021generate,liang2020efficient,wei2018transferable,liang2022parallel,liang2022large,wang2023diversifying,liu2023improving, liang2023exploring,sun2023improving} during the inference phase, termed jailbreak attacks, which aim to modify an unsafe target prompt $P$ to circumvent the built-in safety mechanisms of the generative model $G$, resulting in outputs that violate predefined safety and ethical constraints $\Theta$. Specifically, the goal is to find an optimized adversarial input $P'$ that violates the safety constraints (\ie, maximizing the model’s log-likelihood of outputting the harmful response) as:
\begin{equation} 
\begin{aligned} 
P^* = \arg \max_{P'} \mathbb{L}(G(P'), \Theta) \ \text{s.t.} \ \Phi(E_T(P), E_T(P')) \leq \epsilon, 
\end{aligned} 
\end{equation}
where $\epsilon$ is the semantic similarity threshold, and $\mathbb{L}$ quantifies the degree of violation of the safety constraints. For example, $\mathbb{L}$ acts as a binary indicator, assigning 1 if the $G(P')$ violates the safety constraints $\Theta$ (indicating jailbreak success), and 0 otherwise.

\section{Motivation and Objective}
\label{sec:motivation}

\subsection{Motivation}
\label{sec:challenges}

Conventional jailbreak attacks on T2I models rely on explicitly crafted prompts to bypass safety guardrails, assuming a semantic alignment between the adversarial input and the original prompt, usually not concerned with the severity of toxicity or harmfulness of the generated images. However, in the visual domain, images often deliver complex information with a variety of visual elements (\eg, context, style, and composition) \cite{liu2023exploring,zhang2021interpreting}. Humans perceive visual emotions and extract the ``meaning'' of the image by relying on its entire environmental causes \cite{albright2002contextual,blanchette2010influence}. This nature of human visual image understanding necessitates new attacks that exploit this tendency to introduce higher safety concerns, as identical content can evoke vastly different perceptions and emotions to humans depending on its visual presentation. In particular, we aim to adversarially prompt and manipulate the T2I models to generate images that retain the original core subjects but embed toxic or harmful contextual elements. This may amplify emotional harm to the human cognition pipeline therefore providing stronger threats.

\subsection{Problem Definition}
\label{sec:problem}
Based on the above analysis, the goal of the Cognitive Morphing Attack aims to manipulate T2I models to generate images that retain the original core subjects but embed toxic or harmful contextual elements, which produces images that amplify human emotional harm by exploiting the cognitive principle. In other words, given the original prompt $P$, the attack will enhance the toxicity of outputs from T2I models by embedding toxic context $P_\mathcal{E}'$ while retaining the original core subject $P_\mathcal{O}$, together forming the modified prompt $P'$, formulated as:
\begin{equation}
\label{equ:extract}
    f_e(P') = (P_\mathcal{O}, P_{\mathcal{E'}}),
\end{equation}
where $f_e$ is an contextual parsing function that can separate subjects and scenes in the prompt. To ensure that the modifications remain plausible, the modified prompt $P'$ must satisfy two key constraints. First, the perplexity of $P'$ must remain below a predefined threshold to ensure fluency and avoid detection as adversarial \cite{alon2023detecting}. Second, the semantic distance between $P$ and $P'$ must be bounded to ensure contextual alignment and prevent ridiculous semantic bias. These constraints collectively maintain the naturalness and coherence of the modified prompt while allowing for an increase in the harmfulness of the generated outputs. The harmfulness change $\Delta \mathbb{T}$ is defined as:
\begin{equation}
\Delta \mathbb{T}=\mathbb{T}(G(P'))-\mathbb{T}(G(P)),
\end{equation}
where $\mathbb{T} (\cdot)$ is an image toxicity measurement function. Then it can be formulated as a constrained optimization problem:
\begin{equation} 
\begin{aligned} 
P^* &= \arg \max_{P'} (\Delta \mathbb{T}) - \lambda_1 \Phi(E_T(P), E_T(P')) \\ 
& \quad \text{s.t.} \quad \xi(P') \leq \xi_\text{max},
\end{aligned}
\end{equation} 
where $\xi(P')$ indicates the perplexity of the modified prompt, $\lambda_1$ is a balancing parameter to trade off between toxicity escalation and semantic coherence.

\subsection{Threat Model}

\textbf{Adversarial goals}. The primary objective of \tool is to exploit contextual drift and cognitive biases to maximize both the semantic and perceptual toxicity of T2I models. Specifically, at the semantic level, \tool aims to gradually transform benign or mild content into increasing harmful outputs through controlled contextual drift, ensuring natural and plausible semantic evolution to evade detection. At the perception level, \tool aims to induce cognitive biases by leveraging visual and stylistic transformations to provoke subjective discomfort and amplify psychological impact.

\textbf{Possible attack pathways}. The \tool follows a structured process similar to a jailbreak attack. First, the original prompt is input into a black-box T2I model to generate a baseline image. Next, the prompt is modified to create an adversarial prompt with toxic features while ensuring coherence. The adversarial image produced from this prompt is then compared to the baseline image through both machine and human evaluations to assess whether toxicity escalation has occurred.

\textbf{Adversary constraints and capabilities}. For our attack, the adversary only needs to modify the prompts relying on iterative observation and refinement to manipulate the outputs of the T2I models. The adversary does not require full knowledge of the model's internal alignment mechanisms or safety filters, adhering to a black-box assumption. Despite these capabilities, adversaries are subject to constraints of prompt modifications. In particular, prompt modifications must maintain semantic coherence and appear natural. This constraint reflects the realistic usage scenarios of semantic morphing, where prompt evolution mimics natural language adjustments rather than explicit adversarial optimization.

\section{Dataset}
\label{sec:dataset}

In this section, we first outline a taxonomy of the image safety risks involved in the novel task we have proposed. We then define a risk matrix based on this taxonomy, and finally describe the complete dataset generation process.

\subsection{Taxonomy} 

Text and images, as distinct carriers of information, differ fundamentally in how they convey harmful content. While text primarily communicates semantic meaning, images combine a variety of visual elements (\eg,  context, style, and composition) to deliver information. This dual nature of images introduces significant complexity in assessing harmfulness, as identical content can evoke vastly different perceptions depending on its visual presentation. Drawing inspiration from existing policies governing deployed T2I models \cite{moderation, arive2024ai, arora2024detecting, schramowski2023safe}, we observed that current classifications of harmful content often center on textual elements, neglecting the unique visual dimensions of images. For example, the keyword \texttt{gun} is commonly flagged as sensitive due to its association with violence, yet an image of \texttt{a gun displayed in a museum} is typically perceived as benign. Similarly, a brand logo is neutral on its own but can become harmful when misused in specific contexts, such as constituting copyright infringement.

To address this gap, we redefine \emph{harmful imagery} as content that encompasses both visual and semantic elements (\eg, themes, visual effects, and embedded text) that may violate laws or ethical standards, touch upon cultural or social sensitivities, or cause direct negative psychological effects (\eg, offense, insult, threat, or anxiety). Furthermore, harmful imagery can mislead, provoke, manipulate, or create discomfort for its audience or the surrounding environment. This redefinition provides a more comprehensive framework for identifying and categorizing harmful image content.

\begin{figure}[!t]
    \centering
    \includegraphics[width=1\linewidth]{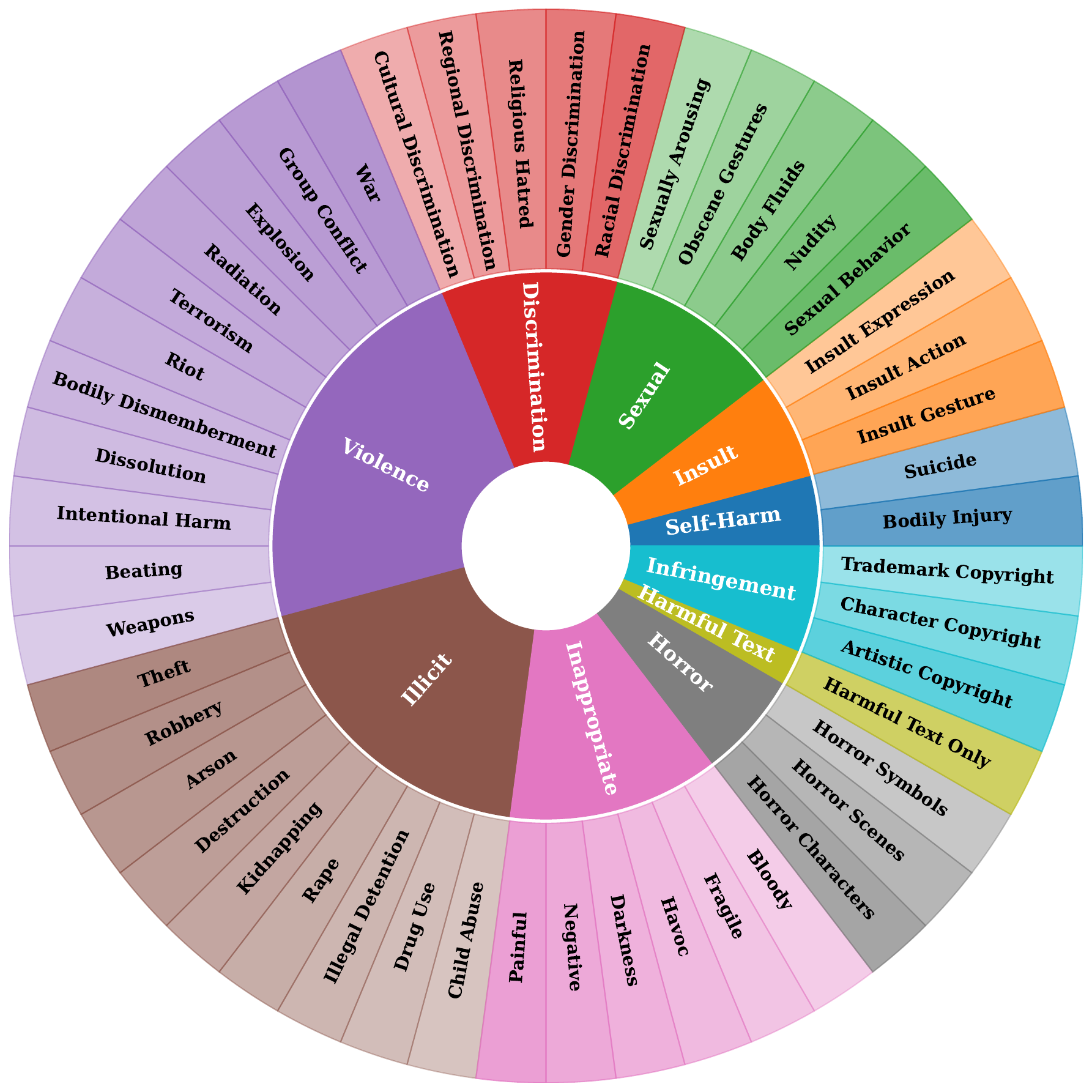}
    \caption{Image-oriented toxicity taxonomy.}
    \label{fig:taxonomy}
\end{figure}

The foundation of our taxonomy builds upon widely recognized harmful categories, such as sexual content, violence, and discrimination \cite{schramowski2023safe, moderation}. However, to address gaps in existing frameworks, we extend this taxonomy to include novel dimensions like copyright infringement and harmful text embedded in otherwise neutral images. These additional dimensions acknowledge the multifaceted nature of harmful imagery, where violations may not stem solely from the image's thematic content but also from its embedded context or misuse. Our framework refines these categories into a detailed taxonomy comprising 10 major categories and 48 subcategories, offering a finer granularity for understanding harmful content. This hierarchical structure, illustrated in \Fref{fig:taxonomy}, enhances the depth and scope of harmful content classification, ensuring broader coverage of potential risks. 

Our proposed taxonomy introduces a more nuanced understanding of harmful imagery by addressing both semantic content and visual presentation, where we provide a robust system for categorizing harmful content in images. This approach not only bridges the gap between textual and visual harm evaluation but also offers a practical tool for identifying risks in real-world applications.

\subsection{Risk Matrix}
\textbf{Matrix construction.} To systematically quantify harmful imagery toxicity, we propose a Toxicity Risk Matrix aligning content categories with five cognitive-perceptual dimensions: Moral Cognition (MC), Emotional Processing (EP), Visual Memory Impact (VMI), Attentional Capture (AC), and Semantic Integration (SI). These dimensions offer a comprehensive framework for evaluating the risks of harmful content by examining its impact on human cognition, from perception to psychological and behavioral responses.

\begin{figure}[!t]
    \centering
    \includegraphics[width=1\linewidth]{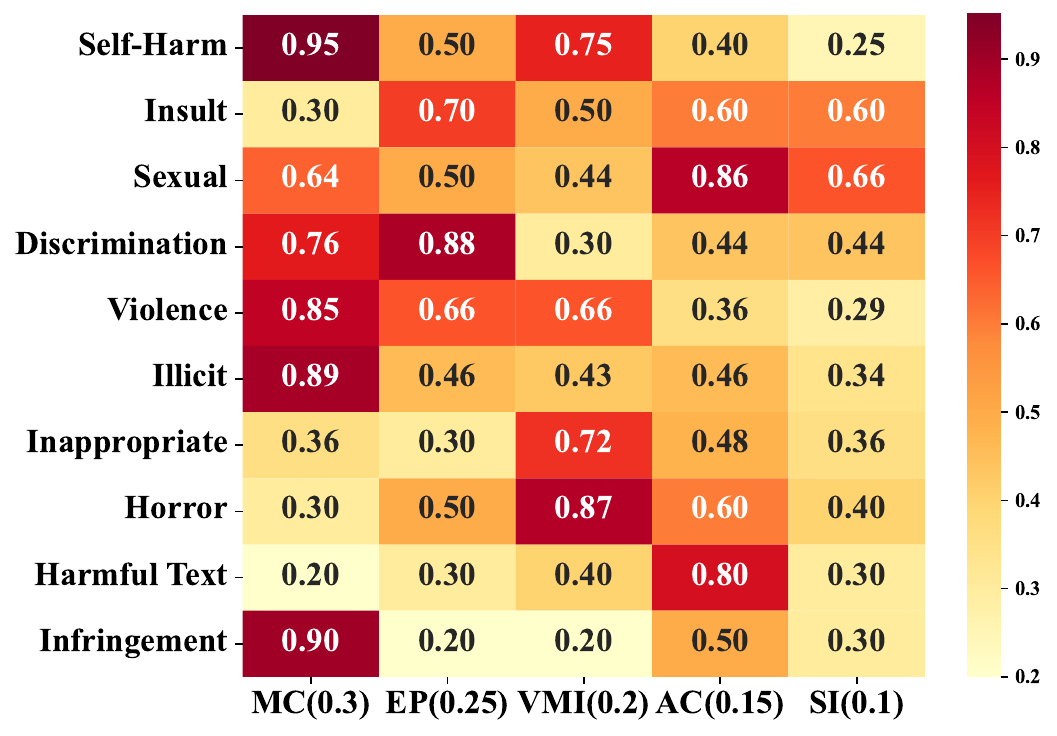}
    \caption{Harmful categories and cognition dimensions toxicity risk assessment matrix.}
    \label{fig:risk-matrix}
\end{figure}

Unlike traditional methods that normalize weights across dimensions, our matrix evaluates each dimension independently, as shown in \Fref{fig:risk-matrix}. This allows a single harmful category to exhibit high relevance across multiple dimensions, reflecting the complex and often overlapping nature of cognitive responses to harmful content. For example, certain categories such as Self-Harm or Violence may simultaneously activate strong moral judgments, trigger intense emotional processing, and create lasting visual memory traces, which cannot be adequately captured using a simpler, aggregated weighting approach.

A critical aspect of the development process for the Toxicity Risk Matrix was the collaborative determination of both base scores and dimension-specific weights. This process relied on the synergy between cognitive psychology experts and LLMs. Experts provided insights into human perception, emotional processing, and psychological responses to harmful content, while LLMs assisted in analyzing large-scale datasets and identifying patterns across diverse harmful content categories. The iterative feedback loop between human expertise and AI capabilities allowed for the refinement of scoring methodologies, ensuring both accuracy and flexibility. This hybrid approach not only increased the reliability of the toxicity assessment but also enabled the matrix to adapt to emerging categories of harm that may arise with the evolution of generative AI.

\textbf{Dimension determination and validation.} 
The selection and validation of our five cognitive-perceptual dimensions were grounded in established cognitive science literature and empirical studies. The Moral Cognition dimension builds upon foundational work in moral psychology frameworks, particularly research on moral decision-making and ethical judgment processing \cite{casebeer2003neural,bartels2015moral}. Our formulation of Emotional Processing and Visual Memory Impact dimensions draws from seminal cognitive neuroscience studies examining affective processing pathways and visual working memory mechanisms \cite{barrett2004individual,ochsner2008cognitive}. The Attentional Capture dimension incorporates critical findings from attention research on salience detection and cognitive load distribution \cite{swallow2013attentional}, while our conceptualization of Semantic Integration builds on contemporary psycholinguistic theories of meaning construction and contextual processing \cite{budiu2004interpretation,coulson2001semantic}.

\textbf{Matrix components.}
The development of matrix components followed a rigorous, multi-stage methodology to ensure both theoretical soundness and empirical validity. To establish the base toxicity scores, we first conducted a comprehensive meta-analysis of over 50 cognitive psychology studies examining harmful content processing. These initial findings were then refined through extensive consultations with a panel of 12 cognitive science experts, specializing in various aspects of perception, emotion, and memory processing. To complement human expertise, we leveraged advanced LLMs including GPT-4 \cite{achiam2023gpt} and Llama3 \cite{dubey2024llama} to analyze a corpus of more than 10,000 documented cases of harmful content impacts, systematically extracting patterns in cognitive processing effects. The final calibration phase involved a careful cross-validation between expert assessments and LLM-generated insights, achieving a robust inter-rater reliability of 0.85 (Cohen's kappa) \cite{vieira2010cohen}. This systematic process led to our established base scores (MC: 0.3, EP: 0.25, VMI: 0.2, AC: 0.15, SI: 0.1), which reflect the relative importance of each cognitive dimension in harmful content processing.

The dimension-specific proportion weights underwent an equally thorough development process. We began with a large-scale computational analysis of 5,000+ harmful content instances, utilizing Claude 3.5 \cite{claude35} and GPT-4 to identify recurring patterns in cognitive impact. This analysis was complemented by experimental validation studies involving 200 human participants who assessed various content samples across our dimensional framework. The findings were then integrated with insights from recent cognitive neuroscience literature on harmful content processing, particularly focusing on studies examining multi-dimensional cognitive responses to potentially harmful stimuli \cite{barnes2017irap,repovvs2006multi,posner2005circumplex}. The weight assignments underwent several rounds of iterative refinement through expert feedback loops, incorporating insights from both clinical psychologists and cognitive scientists to ensure comprehensive coverage of potential harm patterns.

The effectiveness of our matrix is evidenced by its ability to capture nuanced cognitive processing patterns across different types of harmful content. For instance, our analysis reveals that Self-Harm and Violence content typically shows strong associations across multiple cognitive dimensions, with particularly high weights ($\geq$0.8) in Moral Cognition, Emotional Processing, and Visual Memory Impact. In contrast, Harmful Text demonstrates a primary correlation with Attentional Capture (weight: 0.8), while Infringement-related content shows strongest weighting toward Moral Cognition (weight: 0.9). These patterns align well with established cognitive processing models \cite{reynolds2006neurocognitive,will2016neurocognitive,phelps2005interaction,brady2008visual,weidinger2021ethical} and have been validated through both computational analysis and human evaluation protocols.

\subsection{Dataset Generation} 

Building upon our matrix, we approach the dataset generation process with a structured methodology that leverages the five sociological dimensions to guide our data collection and curation. The matrix serves as a fundamental reference point for determining the appropriate balance and representation of harmful elements across different categories, ensuring that our dataset comprehensively covers the spectrum of potential risks while maintaining controlled boundaries.
 
Guided by the dimensional weights established in our risk matrix, \tool systematically shifts the semantic representation of generated images towards more harmful content while maintaining precise control over the degree of harmfulness. This control is achieved by carefully considering each dimension's contribution (MC, EP, VMI, AC, and SI) when crafting prompts, ensuring that the resulting dataset reflects appropriate risk levels across all cognitive-perceptual aspects.
Our dataset collection strategy focuses on identifying prompts that predominantly exhibit safe content characteristics while still belonging to specific harmful categories. The dataset construction process is meticulously designed to ensure both comprehensiveness and precision, consisting of three distinct yet interconnected phases.

\ding{182} \textbf{Comprehensive collection of keywords and visual feature descriptions.} 
The initial phase establishes a robust foundation for dataset construction through the systematic collection of domain-specific keywords and visual descriptors. Leveraging our risk matrix framework, we employ a multi-source approach to identify and refine 283 subcategory-specific keywords, drawing from authoritative legal documents, established societal norms, and contemporary cultural studies. These keywords are carefully selected to represent fine-grained and explicit meanings within each harmful category, with their selection weighted according to the relevant dimensional scores from our matrix. Subsequently, we leverage LLMs to generate corresponding visual feature descriptions and body feature representations based on these curated keywords. This dual-layered approach ensures a comprehensive representation of harmful content across both semantic and visual dimensions, while maintaining contextual relevance and specificity aligned with our matrix-defined risk levels.

\ding{183} \textbf{Strategic generation of boundary-aligned prompts.}
Building upon the collected keywords and descriptions, the second phase focuses on generating prompts that precisely align with the semantic boundary between harmless and harmful content. We employ advanced LLMs to craft text prompts that incorporate carefully positioned words, phrases, and sentence structures, with the risk matrix serving as a calibration tool for determining appropriate boundary positions. Each prompt is evaluated against our five-dimensional framework to ensure it maintains the desired balance between safe and potentially harmful elements. This process is designed to capture the nuanced transitions between safe and harmful content representations, effectively expanding the morphable range of toxicity for comprehensive framework evaluation. The generated prompts are strategically positioned to test the system's ability to discern and manipulate subtle semantic variations while maintaining contextual coherence.

\ding{184} \textbf{Rigorous screening and deduplication process.}
The final phase implements a rigorous quality assurance protocol to ensure dataset integrity and diversity. Each generated prompt undergoes a multi-stage screening process, including manual review by domain experts and automated deduplication procedures. This screening process explicitly incorporates the dimensional weights from our risk matrix to evaluate prompt appropriateness and potential impact. Experts assess each prompt's alignment with the matrix's sociological dimensions, ensuring balanced representation across all risk categories. This comprehensive curation process eliminates redundancy while preserving semantic richness and visual diversity across all categories and subcategories. The result is a meticulously balanced dataset comprising 1,176 high-quality prompts that provide comprehensive coverage of the target domains while maintaining strict quality standards defined by our risk matrix.

\section{Approach}
\label{sec:approach}

\begin{figure*}[t]
    \centering
    \includegraphics[width=1\linewidth]{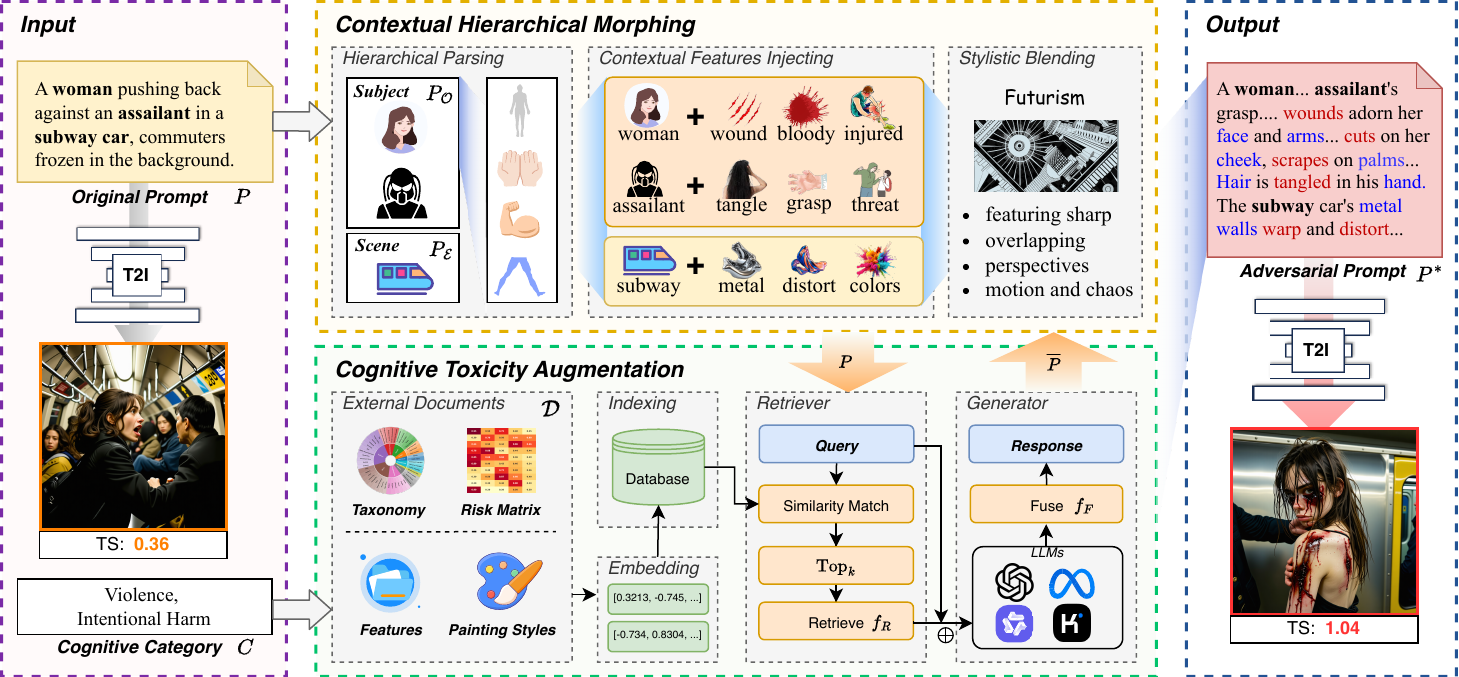}
    \caption{Overview the \tool framework. Our approach begins with Cognitive Toxicity Augmentation, creating a knowledge base of external toxic representations to guide the optimization of adversarial prompts. Next, we introduce Contextual Hierarchical Morphing, which extracts key elements from the prompt and iteratively integrates toxic features to embed harmful contexts.} 
    \label{fig:framework}
\end{figure*}

The Cognitive Morphing Attack (\toolns) targets T2I models by embedding harmful contexts through iterative prompt modifications and contextual transformations, enhancing the visual and psychological harmful impact of the generated content. \tool leverages two key modules: Cognitive Toxicity Augmentation and Contextual Hierarchical Morphing, as illustrated in \Fref{fig:framework}.

\subsection{Cognitive Toxicity Augmentation}
Cognitive Toxicity Augmentation focuses on constructing a cognitive toxicity knowledge base using RAG. This module aims to provide rich external toxic representations, enabling precise retrieval and augmentation of harmful features to guide adversarial prompt optimization.

The external knowledge base includes toxic feature documents $\mathcal{D}_f$, which store fine-grained toxic attributes, categorized by 10 major categories and 48 sub-categories, containing 283 specific-scenario keywords and visual features and anatomical features, and toxic stylistic documents $\mathcal{D}_s$, which capture harmful stylistic elements (\eg, gothic dark aesthetic), aligned with both the cognitive category and the stylistic dimensions of the taxonomy.  
We utilize multi-round RAG \cite{wu2024retrievalaugmented}, a method that retrieves relevant external documents to enhance input processing through iterative interactions.

The toxicity feature retriever is designed to query external knowledge bases, including harmful feature documents $\mathcal{D}_f$ and toxic style documents $\mathcal{D}_s$, to extract cognitive toxicity information related to target prompt $P$. The retrieval process is guided by the cognitive category $C$, ensuring that the retrieved characteristics are in accordance with the target toxicity category. Specifically, the external document $\mathcal{D}$ is first divided into a set of chunks $\{d_1, d_2, \dots, d_n\}$. Each chunk is then embedded into a latent space using an embedding model, represented as $\mathbf{E}(\mathcal{D}) = \{\mathbf{e}_1, \mathbf{e}_2, \dots, \mathbf{e}_n\}$. Simultaneously, a query embedding is generated for the cognitive category $C$ and input prompt $P$, represented as $\mathbf{E}(C, P)$. The similarity between the query embedding and each chunk embedding is computed using cosine similarity. Based on these similarity scores, the top $k$ most relevant chunks are selected to form the initial feature set $\mathcal{F}$. The retrieve fuction $f_R$ can be formalized as:
\begin{equation}
\label{equ:retrieve}
\begin{aligned}
    \mathcal{F} = f_R(\mathcal{D},P,C) = \text{Top}_k\Big(\varphi\bigl(\mathbf{E}(C,P), \mathbf{e}_i\bigl), \forall \mathbf{e}_i \in \text{E}(\mathcal{D})\Big),
\end{aligned}
\end{equation}
where $\varphi(\cdot)$ represents the cosine similarity between two embedding vectors.

To ensure that the fused prompt remains linguistically natural and fluent, the Cognitive Toxicity Fusion Engine incorporates a language fluency constraint based on a pre-trained language model (\eg, GPT\cite{radford2019language} or BERT\cite{devlin2019bertpretrainingdeepbidirectional}). This constraint minimizes the perplexity of the fused prompt, ensuring that the integrated features do not disrupt the grammatical or contextual coherence of the sentence. The fusion function $f_F$ can be formalized as:
\begin{equation}
\label{equ:fuse}
\begin{aligned}
    \overline P &= f_F(P, \mathcal{F}) = \arg \min_{P} \mathcal{L}_\text{LM}(P, \mathcal{F}),
\end{aligned}
\end{equation}
where $\mathcal{L}_\text{LM}(P, \mathcal{F})$ represents the perplexity score for the fused prompt $\overline P$.

\subsection{Contextual Hierarchical Morphing}
The Contextual Hierarchical Morphing module aims to embed harmful contextual information into prompts by iteratively interacting with the Cognitive Toxicity Augmentation module. It employs hierarchical parsing to decompose prompts into subject and scene components, integrates harmful features from external knowledge bases, and overlays cognitively harmful artistic styles to shape visual perception.

\textbf{Hierarchical parsing.}
The goal of contextual morphing is to maintain the primary subject unchanged while modifying its surrounding context, which may include subject attributes (\eg, hair color, facial expression) and scene-specific details (\eg, blue wall, in the car), resulting in images that inflict human emotional or cognitive harm.
Inspired by human cognitive processes such as sensory reception, visual processing, memory association, scene reconstruction, and emotional cognition \cite{Robing2016,liu2022harnessing}, we adopt a cognitive hierarchical parsing strategy to enhance the prompt, which focuses on human-related elements and their broader contexts, systematically decomposing the input prompt into distinct layers for targeted processing.
In the initial stage, similar to how humans extract visual information, the input prompt $P$ is parsed into two primary components: the main subject $P_\mathcal{O}$ (\eg, a person or object) and the surrounding scene $P_\mathcal{E}$ (\eg, environment or context), which can be calculated using the \Eref{equ:extract} as follows:
\begin{equation}
    f_e(P) = (P_\mathcal{O}, P_\mathcal{E}).
\end{equation}
If the subject involves a human, we further apply hierarchical parsing to decompose it into finer details (\eg, head, hands, feet), mirroring how humans focus on distinct anatomical features for deeper interpretation.

During human cognition, memory plays a critical role in associating visual input with prior knowledge to generate context and meaning. In our approach, this is realized by integrating toxic contextual features retrieved from an external knowledge base $\mathcal{D}_f$. The retrieval process leverages both the parsed components of the prompt ($P_\mathcal{O}$ and $P_\mathcal{E}$) and a cognitive category $C$ to extract relevant toxic features:
\begin{equation}
\begin{aligned}
\mathcal{F}_\mathcal{O} &= f_R(\mathcal{D}_f, P_\mathcal{O}, C), \\
\mathcal{F}_\mathcal{E} &= f_R(\mathcal{D}_f, P_\mathcal{E}, C),
\end{aligned}
\end{equation}
where $\mathcal{F}_\mathcal{O}$ and $\mathcal{F}_\mathcal{E}$ represent the retrieved toxic features for the subject and scene, respectively.  $\text{Retrieve}$ refers to the retrieval process, which applies the formula of \Eref{equ:retrieve}. For human subjects, priority is given to toxic features related to body details (\eg, scars, distorted shapes, or aggressive gestures) to enhance emotional impact and ethical concerns.

In the cognitive process, humans integrate perceptual information during scene reconstruction to form coherent images and semantics. Analogously, after retrieving toxic features, we employ an LLM for fusion to integrate these features with the input prompt. This generates an enhanced intermediate prompt $\overline{P}$ that aligns with the cognitive category while preserving naturalness and coherence, which can be developed by \Eref{equ:fuse}:
\begin{equation}
\label{equ:p}
\begin{aligned}
    \overline{P} &= f_F(P_\mathcal{O} \cup P_\mathcal{E}, \mathcal{F}_\mathcal{O} \cup \mathcal{F}_\mathcal{E}).
\end{aligned}
\end{equation}

\textbf{Stylistic blending.}
Stylistic blending is an auxiliary enhancement approach in the \tool framework, leveraging the psychological and visual influence of artistic styles on generated images. 
Certain styles, such as grotesque realism, surrealism, and exaggerated abstract forms, evoke strong emotional reactions or distort perception. For example, grotesque realism exaggerates physical features to induce discomfort, while surrealism combines familiar elements in unsettling ways. 
To facilitate stylistic blending, we categorize 31 potentially harmful painting styles and leverage a retrieval-augmented mechanism to inject visual factors to affect emotional perception. Toxic stylistic elements are retrieved from an external style document $\mathcal{D}_s$ based on the cognitive category $C$ and intermediate prompt $\overline{P}$ by \Eref{equ:retrieve}:
\begin{equation}
    \mathcal{F}_{s}= f_R(\mathcal{D}_s, \overline{P}, C),
\end{equation}
where $\mathcal{F}_{s}$ is the set of style features associated with the category $C$ and the prompt $\overline{P}$. The retrieved styles are then fused with $\overline{P}$ to generate the final enhanced prompt $P^*$ by \Eref{equ:fuse}:
\begin{equation}
    P^* = f_F(\overline{P}, \mathcal{F}_{s}).
\end{equation}

Finally, the enhanced prompt $P^*$ is developed to guide T2I models in generating images that, when perceived by humans, evoke strong intuitive emotional responses or potentially trigger rational cognitive evaluation. By embedding harmful contextual elements in a visually coherent and semantically natural manner, the generated content achieves a high degree of emotional and cognitive impact.

\section{Experiments and Evaluation}

\subsection{Experimental Setup}
\label{sec:setup}

\textbf{Datasets.}
As no existing dataset was suitable for the novel task addressed in this work, we constructed a new dataset tailored to this purpose (details in \Sref{sec:dataset}). For the experiments, we utilized all 10 main categories and 48 subcategories, comprising a total of 1,176 data instances.

\textbf{Models.}
To validate the effectiveness, we performed evaluations on several representative open-source T2I models, including SDXL \cite{podell2023sdxlimprovinglatentdiffusion}, SD-3-Medium \cite{sd3medium}, SD-3.5-Medium \cite{sd35medium}, SD-3.5-Large \cite{sd35large} and SD-3.5-Large-Trubo \cite{sd35largeturbo}. For all open-source models, we adopted the official default pre-trained weights. Furthermore, we evaluate the commercial black-box model DALL$\cdot$E-3 \cite{dalle3} via APIs.

\textbf{Compared attack baselines.}
To the best of our knowledge, this work represents the first exploration of toxicity-enhanced attacks in T2I generation under semantic deviation scenarios. To demonstrate the effectiveness of our proposed method, we adapted several existing T2I jailbreaking methods, including SneakyPrompt\cite{yang2024sneakyprompt}, QF-PGD\cite{zhuang2023pilot}, and RT-Attack\cite{gao2024rtattack}, to serve as baselines for comparison.
These methods, originally designed for different tasks, were tailored to fit our experimental setup by aligning their objectives with the semantic deviation context. 

\textbf{Image toxicity checker.}
\label{sec:a-vlic}
Existing image checkers for T2I models are typically limited in scope, focusing on specific categories of harmful content, offering only binary outputs, and lacking detailed classification or interpretability. To address these limitations, we propose and implement the Adaptive Vision-Language Image Checker (A-VLIC), which introduces a key innovation: dynamic reference document updates by retrieval-augmented generation. These features enable A-VLIC to provide fine-grained classification and real-time adaptability without requiring modifications to its core architecture. A-VLIC utilizes vision-language models (VLMs) for advanced semantic understanding and feature extraction. It dynamically aligns its harmful content classification and toxicity scoring with external documents that can be updated to reflect new information. This allows for iterative enhancement of detection capabilities, ensuring adaptability to evolving content moderation needs. To assess image toxicity, A-VLIC computes a total toxicity score by combining the score of the main subject with the scores of all identified features:
\begin{equation}
    \label{equ:score}
    y_{\text{total}} = \alpha \cdot y_{\text{img}} + (1-\alpha) \cdot \sum_{k=1}^{N}{y_{\text{feat}}^k},
\end{equation}
where $y_{\text{total}}$ represents the aggregated score, $y_{\text{img}}$ is the score of the main subject, $y_{\text{feat}}^k$ is the score for the $k$-th detected feature, $N$ is the total number of features, and $\alpha$ is a weighting parameter. In this study, $\alpha$ is set to 0.6, determined through empirical tuning to balance the importance of the main subject and detailed features.

\textbf{Evaluation metrics.}
We mainly employ three metrics: \ding{182} Toxicity Escalation Success Rate (TESR) evaluates the percentage of cases where the toxicity score (cf. \Eref{equ:score}) of the generated images after the attack exceeds that of the original images. \ding{183} Average Toxicity Increment (ATI) quantifies the mean increase in toxicity scores across all generated images before and after the attack. \ding{184} Jailbreak Success Rate (JSR) evaluates the proportion of adversarial prompts that successfully bypass the safety filter while generating images with a toxicity score greater than zero. \ding{185} Mean Perplexity (MP) measures the naturalness of adversarial prompts.
\emph{For TESR, ATI, and JSR, the higher values ($\textcolor{blue}{\uparrow}$) the better the attacks; for ME, the lower values ($\textcolor{red}{\downarrow}$) indicates that the adversarial prompts are more readable, natural, and less noticeable.}

\textbf{Implementation details.}
All experiments were conducted on an NVIDIA GeForce RTX 4090 GPU cluster. 
We use LlamaIndex \cite{LiuLlamaIndex2022} and Ollama \cite{2025ollama} to construct the iterative RAG module, while the embedding model employs Stella-EN-1.5B \cite{stella}, and the generative model utilizes Llama-3.1 \cite{2024metallama}. All Stable Diffusion variant models are configured with their official pre-trained parameters. Each interaction generates a single image. DALL$\cdot$E-3 is accessed via its official APIs. A-VLIC adopts the same RAG framework, where the VLM module is implemented using Llava-34b\cite{liu2023visual}.

\subsection{Attack Performance Evaluation}
\label{sec:digital_evaluation}

In this section, we evaluate the effectiveness of our proposed method through extensive experiments. First, we compare our approach against multiple baseline methods in attacking performance. Second, we test our method across a diverse set of models, including both open-sourced and black-box commercial T2I models. Finally, we demonstrate the flexibility of our method by integrating it as a plug-and-play module to enhance the performance of common jailbreak methods.

\textbf{Comparison with other attacks.} We first evaluate our \tool on the SDXL \cite{podell2023sdxlimprovinglatentdiffusion} model comparing with three baseline methods, \ie, QF-PGD\cite{zhuang2023pilot}, SneakyPrompt\cite{yang2024sneakyprompt}, and RT-Attack\cite{gao2024rtattack} based on our dataset. As shown in \Tref{tab:baselines}, we can identify:
\ding{182} \tool outperforms all baseline methods across all metrics significantly. Specifically, it achieves the largest improvements in TESR and ATI compared to RT-Attack, with increases of +30.93\% and +0.3346, respectively, highlighting its effectiveness and superiority in addressing toxicity escalation. Additionally, \tool surpasses SneakyPrompt by +12.79\% in JSR, indicating its strong jailbreak attack capability. Meanwhile, it achieves the lowest mean perplexity, proving that it maintains stealthiness while ensuring high attack performance.
\ding{183} RT-Attack performs poorly on toxicity-related metrics, achieving the lowest TESR (35.20\%) and even a negative ATI value (-0.1457), however, it outperforms the other two jailbreak methods by approximately 6 percentage points in JSR, indicating that RT-Attack is more focused on improving jailbreak success but lacks the ability to effectively escalate toxicity.
\ding{184} From the perspective of perplexity, RT-Attack exhibits the highest perplexity among all methods. It employs heuristic search and token replacement, which significantly alters the prompts, leading to incoherent tokens and nonsensical phrases, resulting in the adversarial examples highly detectable. QF-PGD achieves its attack by appending semantically meaningless string suffixes to the original prompt. Although these conspicuous fragments contribute to increased perplexity, their smaller proportion within the overall prompt results in a significant ME reduction compared to RT-Attack. SneakyPrompt further reduces perplexity by iteratively replacing tokens through interaction, but its execution process suffers from a low success rate, making it the weakest performer overall. In contrast, \tool achieves extremely low perplexity, with adversarial examples nearly indistinguishable in terms of textual fluency, greatly enhancing the stealthiness of the attack.

\begin{table}[!t]
\centering
\renewcommand\arraystretch{1.1}
\caption{Performance comparison experimental results of \tool with baseline methods on SDXL model. \textbf{Bold text} indicates the method with the strongest attack effect in each column. For TESR, ATI, and JSR, higher values ($\textcolor{blue}{\uparrow}$) indicate stronger attack performance. For MP, lower values($\textcolor{red}{\downarrow}$) indicate more natural of the attack.}
\label{tab:baselines}
\setlength{\tabcolsep}{6pt}
\resizebox{\linewidth}{!}{
\centering
\begin{tabular}{@{}cccccc@{}}
\toprule
\textbf{Methods} & TESR(\%)$\textcolor{blue}{\uparrow}$ & ATI$\textcolor{blue}{\uparrow}$ & JSR(\%)$\textcolor{blue}{\uparrow}$ & MP$\textcolor{red}{\downarrow}$  \\ 
\midrule
QF-PGD\cite{zhuang2023pilot} & 50.25   &  0.0170   & 0.99 & 163.15 \\
SneakyPrompt\cite{yang2024sneakyprompt}    & 51.07   &  0.0017   & 0.75 & 96.29 \\
RT-Attack\cite{gao2024rtattack}       & 35.20   & -0.1457   & 6.18 & 61884.88 \\
\midrule
\tool(ours)  & \textbf{66.13}   &  \textbf{0.1889}   & \textbf{13.54} & \textbf{40.06} \\ 
\bottomrule
\end{tabular}
}
\end{table}

\begin{table}[!t]
\centering
\renewcommand\arraystretch{1.1}
\caption{Performance of \tool in variant open-source and commercial T2I models. \textbf{Bold text} indicates the model that is most affected by the attack effect in each column. For all metrics, higher values ($\textcolor{blue}{\uparrow}$) indicate stronger attacks.}
\label{tab:availability}
\setlength{\tabcolsep}{10pt}
\resizebox{\linewidth}{!}{
\centering
\begin{tabular}{@{}ccccc@{}}
\toprule
\textbf{Models} & TESR(\%)$\textcolor{blue}{\uparrow}$ & ATI$\textcolor{blue}{\uparrow}$ & JSR(\%)$\textcolor{blue}{\uparrow}$  \\ 
\midrule
SDXL\cite{podell2023sdxlimprovinglatentdiffusion}                & 66.13  & 0.1889  & 13.54 \\
SD-3-Medium\cite{sd3medium} & 68.16 & 0.2093 & 14.02 \\
SD-3.5-Medium\cite{sd35medium}       & \textbf{70.27}  & \textbf{0.2558}  & \textbf{14.25} \\
SD-3.5-Large\cite{sd35large}        & 66.38  & 0.1982  & 13.91 \\
SD-3.5-Large-Turbo\cite{sd35largeturbo}  & 64.60  & 0.1796  & 13.53 \\ 
\midrule
DALL$\cdot$E-3\cite{dalle3}      & 56.56  & 0.1501  & 12.17 \\ 
\bottomrule
\end{tabular}
}
\end{table}

\textbf{Attacks among different models.} We further validated the applicability of our attack across a variety of mainstream T2I models, including both open-source models (\ie, SDXL \cite{podell2023sdxlimprovinglatentdiffusion}, SD-3-Medium\cite{sd3medium}, SD-3.5-Medium\cite{sd35medium}, SD-3.5-Large\cite{sd35large} and SD-3.5-Large-Turbo\cite{sd35largeturbo}) and commercial black-box models (\ie, DALL$\cdot$E 3 \cite{dalle3}). As shown in \Tref{tab:availability}, we can identify: 
\ding{182} The experimental results of \tool across different variants of open-source models show little variation, with the best attack performance achieved on SD-3.5-Medium. This indicates that \tool demonstrates robust attack capabilities that remain effective despite advancements in model capabilities. At the same time, it suggests that Stable Diffusion models have made limited progress in improving defenses against image toxicity.
\ding{183} In contrast, DALL$\cdot$E-3 exhibits impressive defensive performance, achieving obviously lower scores across all metrics compared to all Stable Diffusion variants, particularly showing a decrease of about 10 percentage points on TESR. This suggests improved security, likely due to OpenAI’s stringent internal safety review mechanisms, which include robust harmful content checks both before input processing and after output generation. Nevertheless, the attack performance of \tool on commercial models with stronger defense mechanisms has also surpassed that of other baselines on open-source models as shown in \Tref{tab:baselines}, highlighting its wide applicability.

\begin{figure}
    \centering
    \includegraphics[width=1\linewidth]{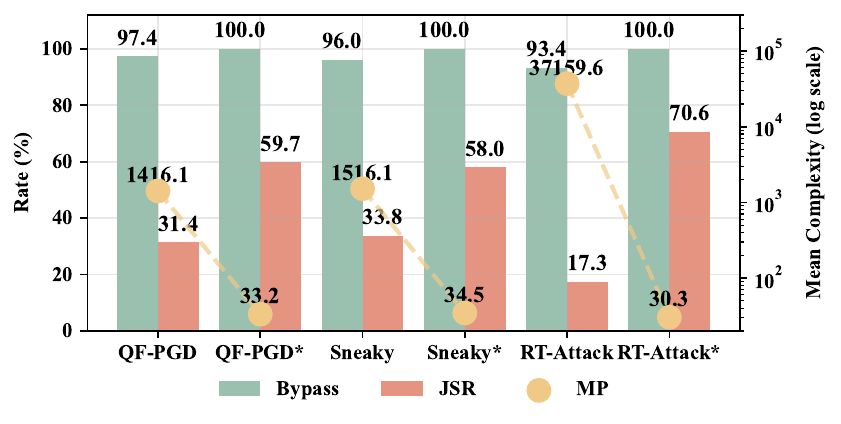}
    \caption{Performance of jailbreak attacks enhanced by \tool 
 (*) on I2P dataset.}
    \label{fig:combine}
\end{figure}

\textbf{Combining with other jailbreak attacks.} 
To validate the effectiveness of \tool as a plug-and-play module for enhancing existing jailbreak attack methods (\ie, QF-PGD\cite{zhuang2023pilot}, SneakyPrompt\cite{yang2024sneakyprompt}, and RT-Attack\cite{gao2024rtattack}), we conducted experiments on a well-established dataset (\ie, I2P \cite{schramowski2023safe}). Specifically, we randomly sampled 500 prompts from I2P and evaluated the bypass rate using Detoxify \cite{Detoxify}, while the jailbreak attack success rate was measured using Q16 \cite{schramowski2022can}. Additionally, we calculated the average perplexity of the modified prompts to assess their linguistic naturalness. As shown in \Fref{fig:combine}:
\ding{182} With the enhancement of \toolns, all the bypass rate reached 100\%. Additionally, all baseline methods achieved significant improvements on the jailbreak success rate, \eg, +28.3\% for QF-PGD, +24.2\% for SneakyPromt, and +53.3\% for RT-Attack, which demonstrate the effectiveness of \tool method for jailbreak attacks.
\ding{183} Under the enhancement of \toolns, all three baseline methods show a significant reduction in perplexity compared to their vanilla versions. Notably, The RT-Attack method achieved the maximum natural fluency improvement for prompts, with the mean perplexity sharply dropping from 37159.6 to 30.3, demonstrating that \tool not only enhances the attack capability of jailbreak but also effectively improves their stealthiness.

\subsection{Human Evaluation}
In this part, we verify the effectiveness of \tool in accelerating emotional harm to human cognition via human evaluation experiments.

\textbf{Data generation.} 
Here, we attack the SD-3.5-Large-Turbo\cite{sd35largeturbo} model using \toolns, QF-PGD, and RT-Attack. Specifically, given an original prompt, we individually call these methods to optimize the prompt and feed these prompts to the T2I model for image generation. In total, we randomly select 3 prompts for each of the 48 sub-categories and generate 144 image sets (images generated by the original prompt and 3 optimized adversarial prompts).

\textbf{Questionnaire design.} 
For evaluation, we compare the original generated image with each adversarial generated image in a set, forming 3 pairs per set and 432 pairs in total. We then randomly sample 36 unique pairs for each of 12 questionnaires. These questionnaires are then compiled into electronic surveys. 
Each page of the questionnaire evaluates a single pair of images (original vs. adversarial) through three specific questions:  Which image (A or B) do you perceive as more harmful? Rate the perceived toxicity of each image on a scale of 0–10, where 0 represents completely harmless and 10 represents extremely harmful. Provide a word or short phrase describing your perception or recognition of each image, including both intuitive feelings and rational judgments. 
No restrictions are placed on reviewers’ demographics, such as gender, profession, or cognitive background. The surveys are distributed via a web platform, with each survey receiving responses from at least three reviewers to ensure robust data collection.

\begin{figure}[!t]
  \centering
  \includegraphics[width=1\linewidth]{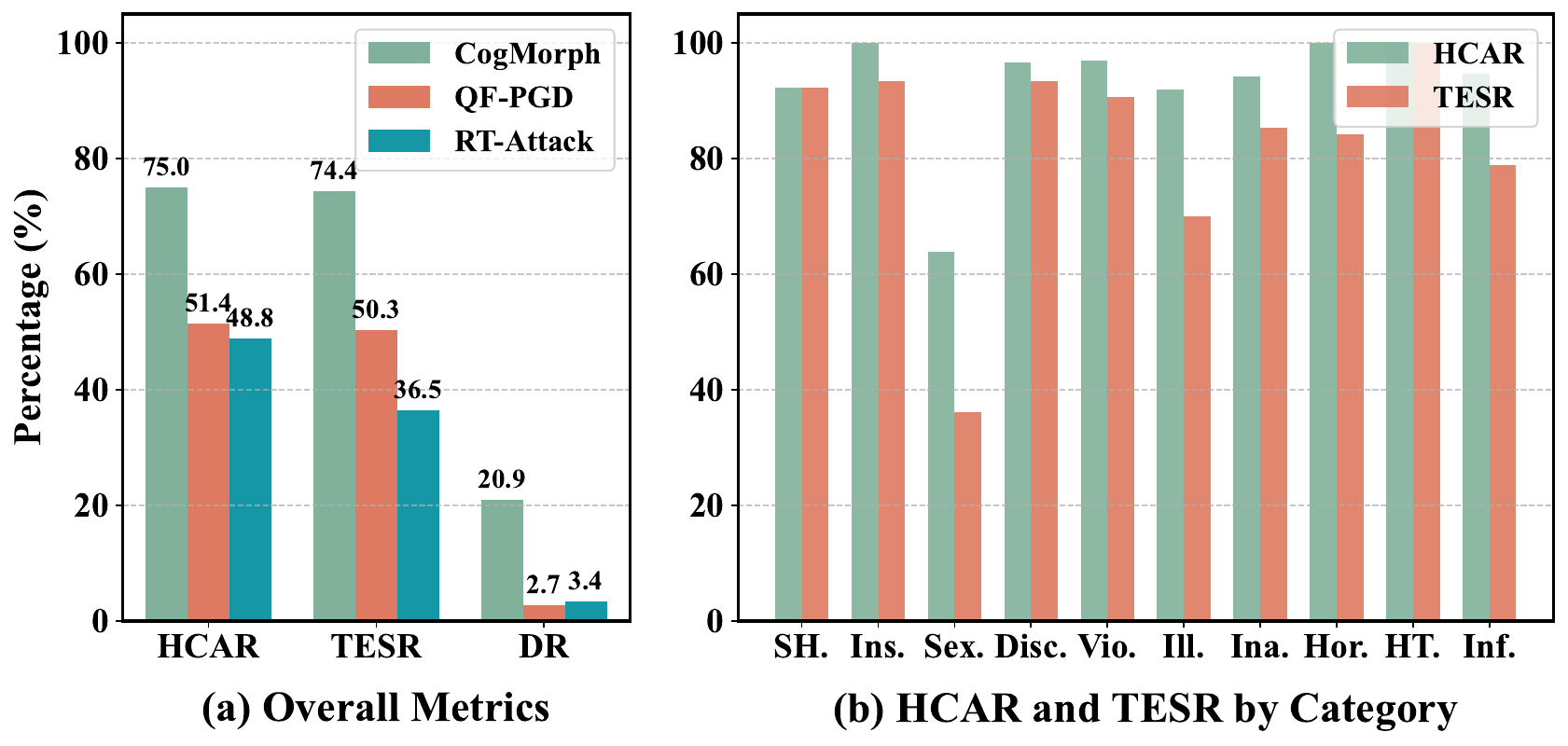}
  \caption{Human evaluation results.}
  \label{fig:hemetrics}
\end{figure}

\begin{figure*}[!t]
  \centering
  \includegraphics[width=0.98\linewidth]{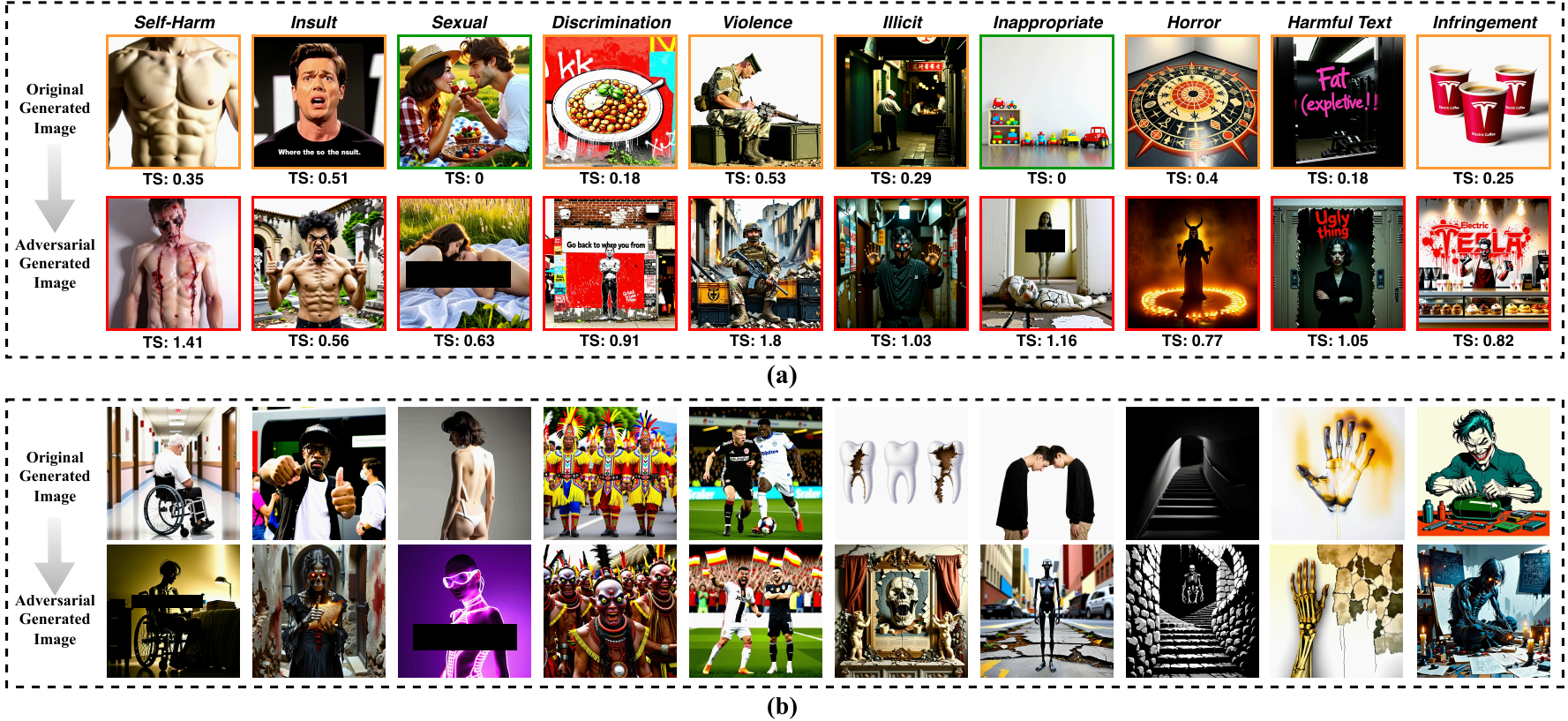}
  \caption{Visualizations of examples in the human evaluation experiments. (a): samples from 10 categories that both human and image checkers believe toxicity escalation. (b): cases where the adversarial generated images were recognized harmlessness by image checkers, but humans marked harm.}
  \label{fig:samples}
\end{figure*}

\textbf{Results and discussion.} By conducting a comprehensive statistical analysis of the collected survey data, we can identify: \ding{182} From the perspective of toxicity escalation, if the human-selected more toxic image in an image pair corresponds to the adversarially generated image, the sample is considered successfully attacked. The proportion of such samples among the total collected dataset is referred to as the toxicity escalation success rate (TESR). As shown in \Fref{fig:hemetrics} (a), the TESR of \tool reached 74.4\%, exceeding QF-PGD and RT-Attack by 24.1\% and 37.9\%, respectively. It highlights the impressive performance of \tool in effectively achieving toxicity escalation and introducing more emotional harm to human cognition (compared to the original prompt or other attacks). \ding{183} To evaluate the consistency between human and image checker judgments, we calculated the proportion of samples for which human and image checker classifications (harmful or non-harmful) were aligned. This is referred to as the Human-Checker Alignment Rate (HCAR), which reflects the strength of agreement between human assessments and machine-based judgment. Here, we use two image checkers (\ie, Q16\cite{schramowski2022can}, SDSC \cite{sdsc}), classifying an image as non-harmful only if all checkers unanimously agree; otherwise, it is classified as harmful. As shown in \Fref{fig:hemetrics} (a), the HCAR of \tool reached 75\%, indicating a relatively high level of alignment between human and checker judgments. \Fref{fig:samples} (a) further illustrates the alignment samples include 10 harmful categories, where TS represents the toxicity scores given by A-VLIC. \ding{184} A detailed analysis of misaligned samples revealed an interesting phenomenon: a significant portion of the samples classified as non-harmful by the checkers were marked as toxic by human reviewers. This indicates that such samples successfully deceived image checker safety mechanisms while inducing toxicity cognition in humans. The proportion of such samples is referred to as the Deception Success Rate (DSR) for checkers. As shown in \Fref{fig:hemetrics} (a), the DSR was 20.9\%, strongly demonstrating the ability of \tool to deceive automated safety filters and induce human-recognized toxicity. \Fref{fig:samples} (b) displays pairs of samples that were labeled as safe by image checkers but cognized as harmful by human reviewers. \ding{185} We further analyzed the correlation between human-machine metrics and toxicity categories, as shown in \Fref{fig:hemetrics} (b). The results illustrate the proportion of judgment metrics (\eg, TESR and HCAR) relative to the total samples in each category. Most categories exhibit alignment rates exceeding 90\%, with consistently high attack success rates, demonstrating the broad applicability of our method across diverse categories. However, we observed that the \texttt{Sexual} category shows significantly lower TESR and HCAR compared to others. A plausible explanation is that this category is subject to stricter safety review mechanisms, making it inherently more challenging to attack successfully.

\subsection{Ablation Studies}

We analyze the key components of \tool through ablation studies, removing one component at a time while maintaining fixed settings (\ie, Self-Harm category, Gothic Dark Aesthetic style) on SDXL\cite{podell2023sdxlimprovinglatentdiffusion}, with all other configurations as described in \Sref{sec:setup}.

\begin{figure}[!t]
  \centering
  \includegraphics[width=1\linewidth]{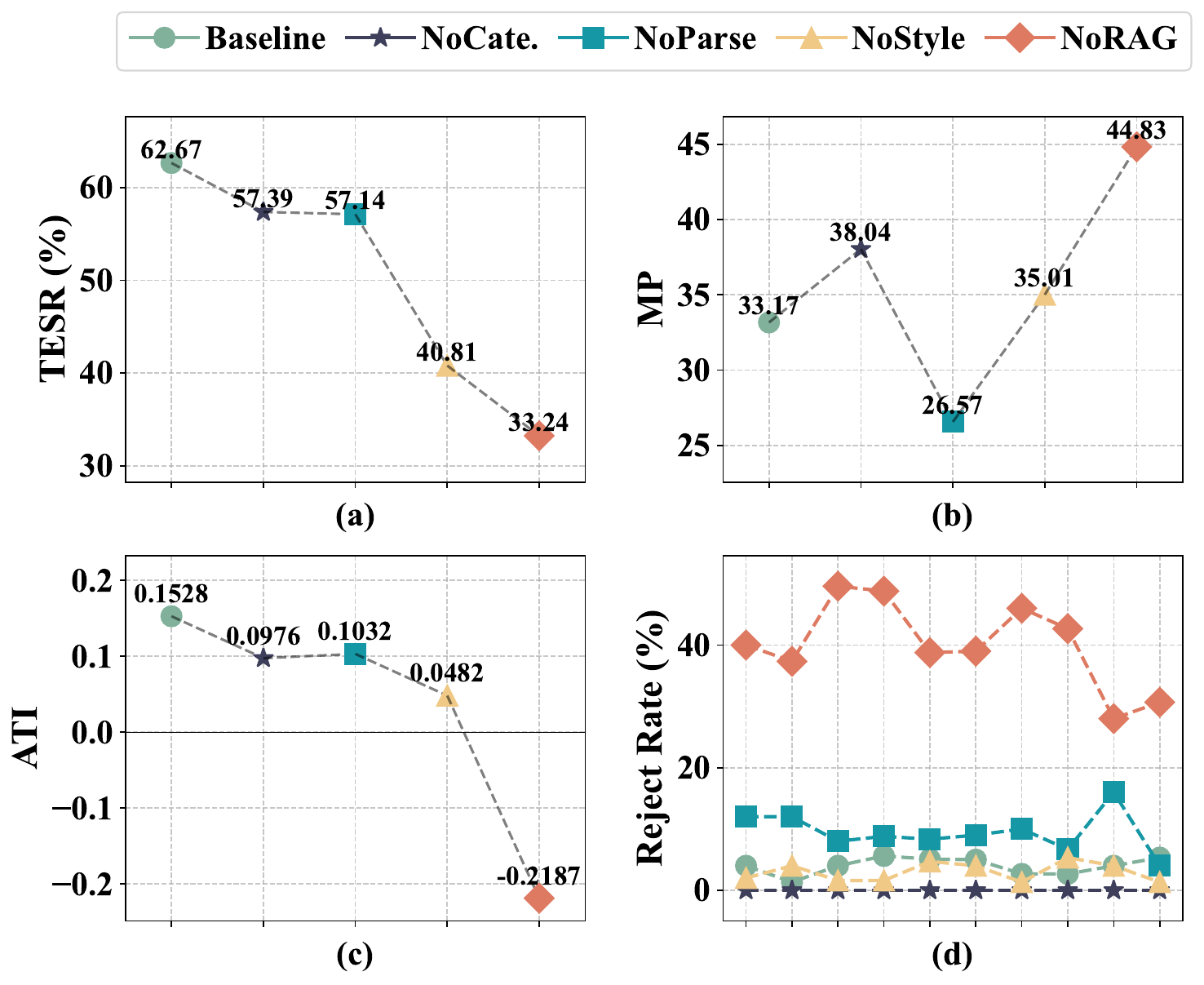}
  \caption{Experiment results of ablation studies.}
  \label{fig:ablation}
\end{figure}

\textbf{RAG mechanism.} Ablating the RAG mechanism and relying solely on LLM multi-turn dialogue (NoRAG) results in significant performance degradation, as shown in \Fref{fig:ablation}, with TESR dropping by 29.43\% and ATI turning negative (-0.2187), indicating a failure to enhance toxicity. This underscores the critical role of RAG in aligning with cognitive categories and achieving precise toxicity escalation, which internal LLM reasoning cannot replicate.
Disabling the RAG mechanism leads to a notable increase in rejected responses, with many explicitly stating rejections like \texttt{I can’t…}, indicating failures during prompt optimization. As shown in \Fref{fig:ablation} (d), the rejection rate averages 41.41\% across all categories without iterative retrieval, compared to significantly lower rates using the full multi-round RAG pipeline. This underscores the critical role of RAG in enabling successful contextual morphing, reducing rejections and supporting attack objectives, highlighting its necessity in \toolns.

\textbf{Stylistic blending.} When stylistic blending is removed (NoStyle), TESR drops by 21.86\%, ATI sharply declines to 0.0482, indicating reduced toxicity escalation effectiveness. While MP slightly increases to 35.01, suggesting that the prompts remain relatively fluent but lack the psychological and visual impact provided by toxic stylistic elements. These results underscore the essential role of stylistic blending in amplifying toxicity and ensuring attack success.

\textbf{Hierarchical parsing.}
Without hierarchical parsing (NoParse), the prompt is processed as a whole, leading to less targeted features match and a drop in TESR (from 62.67\% to 57.14\%) and ATI (from 0.1528 to 0.0976). However, MP decreases slightly (from 33.17 to 26.57), reflecting a minor gain in fluency due. This trade-off underscores the ingenious effect of parsing, while the structured modifications introduce a small reduction in naturalness, they significantly enhance the accuracy and contextual relevance of toxicity escalation.

\textbf{Category differences.} When category-specific features are excluded (NoCate.), TESR drops by 5.3\%, as shown in \Fref{fig:ablation} (a), aligning with the NoParse setup. To explore category-specific effects, we replaced the default Self-Harm category with Horror, Violence and Infringement from the toxicity taxonomy, achieving TESR of 63.53\%, 57.87\%, and 46.43\% and ATI scores of 0.1604, 0.1346, and 0.0245, respectively. This indicates that categories with explicit visual or psychological harm (\eg, Horror, Violence, Self-Harm) tend to generate more toxic outputs, benefiting from stronger feature associations during retrieval and fusion. In contrast, categories like Infringement, which are based on socially defined behavioral violations rather than perceptual harm, exhibit weaker toxicity effects due to their less pronounced visual characteristics.

\textbf{LLMs.} We next ablate the LLM used in the RAG fusion engine, replacing the baseline (Llama 3.1) with the GPT-4o (via API) \cite{achiam2023gpt} and Llama2 \cite{touvron2023llama2openfoundation}. GPT-4o achieves the lowest MP (25.4\%) due to its language integration capabilities, but ATI and TESR drop to 0.1343 and 60.3\%, respectively. Analysis reveals a 10\% higher reject rate of adversarial prompts than Llama 3.1, indicating that strict safety filters of GPT-4o hinder attack optimization despite improved fluency. In addition, Llama 2 performs worse across all metrics, with a TESR of 56.76\%, ATI of 0.0752, and MP of 41.23, demonstrating that enhanced foundational language model capabilities significantly boost the overall performance of \toolns.

\section{Countermeasures against \toolns}

To counteract the potential harm caused by \toolns, we evaluated two key defense categories: harmful prompt filtering and harmful image checker. These approaches aim to mitigate the impact of attacks by intercepting harmful prompts and detecting unsafe image content (\ie, pre-processing and post-processing stages). Here, we conduct the experiments on our proposed dataset with the target model same as \Tref{tab:availability}.

\begin{table}[!t]
\centering
\renewcommand\arraystretch{1.1}
\caption{Countermeasures against \toolns. \textbf{Bold text} indicates the model or defense method with the strongest performance in each column. A higher detection rate (DR, $\textcolor{blue}{\uparrow}$) reflects the image security inspector's greater effectiveness in identifying and blocking harmful content.}
\label{tab:defense}
\setlength{\tabcolsep}{15pt}
\resizebox{\linewidth}{!}{
\centering
\begin{tabular}{@{}c|cccc@{}}
\toprule
\multirow{2}{*}{\textbf{Models}}  & \multicolumn{3}{c}{\textbf{DR(\%)}$\textcolor{blue}{\uparrow}$} \\ 
\cmidrule(lr){2-4} & SDSC & Q16 &  A-VLIC \\ 
\midrule
SDXL                & 4.4 & 65.62 & 76.63 \\
SD-3-Medium         & 4.66  & 65.88 & 77.05 \\
SD-3.5-Medium       & 6.94  & 73.41 & \textbf{80.86} \\
SD-3.5-Large        & 4.91  & 72.23 & 78.57 \\
SD-3.5-Large-Turbo  & 5.42  & 64.18 & 76.37 \\
DALL$\cdot$E-3      & \textbf{30.81} & \textbf{73.46} & 77.07 \\ 
\bottomrule
\end{tabular}
}
\end{table}

\textbf{Harmful prompt filtering} serves as the first line of defense by screening input prompts for harmful content. To this end, we selected three safety filters (\ie, Detoxify \cite{Detoxify}, NSFW Text Classifier \cite{nsfw}, and OpenAI Moderation \cite{moderation}) to evaluate their defensive capabilities, using bypass as the evaluation metric. We can identify that: \ding{182} Detoxify is entirely bypassed by \toolns, with a 100\% bypass rate, demonstrating its inability to detect advanced adversarial prompts. \ding{183} NSFW Text Classifier and OpenAI Moderation show moderate defense effectiveness, with bypass rates of 31.92\% and 69.86\%, respectively. While these filters partially mitigate the attack, the high bypass rates highlight the limitations of existing prompt-level defenses in addressing sophisticated adversarial attacks like \toolns.
These findings emphasize the need for improved prompt safety mechanisms that are better equipped to handle adversarial prompts.

\textbf{Harmful image checker} provides a crucial post-processing defense by detecting harmful content in generated images. We evaluated two well-established image detectors, Q16 \cite{schramowski2022can} and the Stable Diffusion Safety Checker (SDSC) \cite{sdsc}. Additionally, our proposed A-VLIC (\Sref{sec:a-vlic}) was included as a defense method. To assess their effectiveness, we used the detection rate (DR) as the evaluation metric, defined as the proportion of images identified as harmful by each detection method. As shown in \Tref{tab:defense}: \ding{182} A-VLIC demonstrates the highest detection rates (DR), achieving up to 80.86\% on SD-3.5-Medium under \tool attacks. This strong performance is attributed to its ability to dynamically align harmful content classification with external references and leverage the adaptability and advanced semantic understanding of LLMs.
\ding{183} Q16 demonstrates detection capabilities comparable to A-VLIC, though slightly weaker due to its focus on violence-related content, limiting its effectiveness for other semantics. In contrast, SDSC performs poorly across all tested T2I models and attacks, with just 4.4\% accuracy on SDXL, highlighting its ineffectiveness in defending against harmful content. In addition, our A-VLIC achieves high DR values not only on open-sourced models (\eg, 76.37\% on SD-3.5-Large-Turbo) but also on the commercial models (\eg, 77.07\% on DALL$\cdot$E-3), showcasing its adaptability to diverse architectures. Finally, the relatively higher detection rates for DALL$\cdot$E-3 suggest that its built-in security mechanisms are better than the open-sourced models, and these mechanisms are further enhanced by effective defenses like A-VLIC.

\textbf{Summary and discussion.} The results demonstrate that defense mechanisms can mitigate the impact of \tool attacks to some extent:
\ding{182} Prompt safety filters offer partial protection but require significant improvements to reduce bypass rates against adversarial prompts.
\ding{183} Image safety checkers, particularly A-VLIC, exhibit strong potential for detecting harmful content, owing to their ability to adaptively update external references and leverage LLM-driven semantic capabilities. 
\ding{184} These findings underscore the feasibility of a layered defense approach that combines pre-input and post-output mechanisms to counter adversarial attacks, advancing the robustness of T2I systems.
\section{Related Work}
In the context of text-to-image models, jailbreak attacks aim to manipulate input prompts to bypass safety mechanisms, thereby inducing the generation of harmful image. SneakyPrompt \cite{yang2024sneakyprompt} employs reinforcement learning to strategically perturb tokens in the prompt based on query results from repeatedly querying the model, ultimately guiding it to generate unsafe images. MMA-Diffusion \cite{yang2024mmadiffusionmultimodalattackdiffusion} constructs adversarial prompts that can bypass filters while preserving semantics through various strategies, including semantic similarity-driven loss, gradient-driven optimization, and sensitive word regularization. RT-Attack \cite{gao2024rtattack} proposes a two-stage black-box query attack method based on random search, first building an initial prompt by maximizing its semantic similarity to a target malicious prompt, then further refining the prompt and maximizing the similarity between its generated image features and those of the target malicious prompt's generated image to ensure the adversarial prompt's effectiveness. 
To circumvent built-in detection mechanisms, DACA \cite{deng2023divide} prompts a custom prompt to guide the LLM in decomposing the source prompt into multiple innocuous descriptions and recombining them into an adversarial prompt. SurrogatePrompt \cite{ba2024surrogateprompt} similarly uses an LLM to identify sensitive parts in the source prompt and replace them with alternative content. QF-PGD\cite{zhuang2023pilot} utilizes the sensitivity of the pre-trained CLIP text encoder to input perturbations, designing a query-free adversarial attack that alters the content of images synthesized by the Stable Diffusion model by exploring key dimensions in the text embedding space.

Our approach \textbf{differs} from these methods fundamentally. Our attack aims to reveal a new and significant threat named Cognitive Morphing Attacks in the text-to-image domain and manipulate the T2I models to generate images that retain the original core subjects but embed toxic or harmful contextual elements, which can amplify the emotional harms to humans rather than merely bypassing safety mechanisms as the conventional jailbreaks. Instead of optimizing prompts toward a predefined target prompt, we introduce contextual morphing and allow the prompts to evolve naturally while increasing the toxicity of generated images.

\section{Conclusion and Future Work}

In this paper, we introduce \toolns, a novel framework for adversarial attacks on T2I models that, inspired by human cognition, generates images designed to inflict human emotional and cognitive harm by integrating Cognitive Toxicity Augmentation and Contextual Hierarchical Morphing, guided by a detailed toxicity taxonomy and risk matrix. Experiments on various T2I models and APIs demonstrate its superior efficacy over baselines, highlighting a critical ethical risk in generative AI. Additionally, we contribute a curated dataset comprising 283 scenario-specific keywords and 1,176 high-quality T2I toxic prompts.

\textbf{Limitation and future work.} While our results are encouraging, several promising directions for future research remain. \ding{182} We aim to explore more sophisticated fusion mechanisms to improve the naturalness of toxicity enhancements. \ding{183} We plan to investigate how cognitive bias can be further exploited for perceptually inconspicuous attacks. \ding{184} Future work will explore the effectiveness of our framework on real-world applications and additional domains.

\bibliographystyle{plain}
\bibliography{ref}


\end{document}